# Data Augmentation through Background Removal for Apple Leaf Disease Classification Using the MobileNetV2 Model

Youcef Ferdi
National Higher School of iotechnology
Constantine, Algeria
y.ferdi@ensbiotech.edu.dz

*Abstract*—The advances in computer vision made possible by deep learning technology are increasingly being used in precision agriculture to automate the detection and classification of plant diseases. Symptoms of plant diseases are often seen on their leaves. The leaf images in existing datasets have been collected either under controlled conditions or in the field. The majority of previous studies have focused on identifying leaf diseases using images captured in controlled laboratory settings, often achieving high performance. However, methods aimed at detecting and classifying leaf diseases in field images have generally exhibited lower performance. The objective of this study is to evaluate the impact of a data augmentation approach that involves removing complex backgrounds from leaf images on the classification performance of apple leaf diseases in images captured under real world conditions. To achieve this objective, the lightweight pre-trained MobileNetV2 deep learning model was fine-tuned and subsequently used to evaluate the impact of expanding the training dataset with background-removed images on classification performance. Experimental results show that this augmentation strategy enhances classification accuracy. Specifically, using the Adam optimizer, the proposed method achieved a classification accuracy of 98.71% on the Plant Pathology database, representing an approximately 3% improvement and outperforming state-of-the-art methods. This demonstrates the effectiveness of background removal as a data augmentation technique for improving the robustness of disease classification models in real-world conditions.

## I. INTRODUCTION

Plant disease detection and classification methods are of paramount importance for farmers as they assist them in identifying and addressing diseases early, thereby minimizing crop losses and increasing yields. While traditional methods are still used in many places, nowadays, there is a growing trend of using innovative technological methods in agriculture, including the use of smartphone apps and remote sensing technologies for faster and more accurate disease detection and identification.

Machine learning algorithms and deep learning models, often combined with image processing techniques, play crucial role in the development of increasingly accurate methods for the detection and classification of plant diseases by analyzing images of diseased plant organs. In plant disease classification, leaf images are among the most commonly used organ images. Leaves are frequently the most visibly impacted parts of a plant by diseases, making them the primary focus for disease detection and classification algorithms.

In recent years, deep learning models, particularly convolutional neural networks (CNNs), have been successfully applied in the agriculture domain [1, 2, 3]. Many datasets of plant leaf images are publicly available on the internet such as Plant Village [4], and Plant Pathology [5]. Images of plant diseases available in these datasets are captured using various methods and devices, either in a laboratory [4] or in the field [5]. In a laboratory setting, controlled conditions like background, lighting, temperature, etc., may be employed to ensure a high-quality image acquisition. On the other hand, images captured in the field reflect the real-world conditions including variations in lighting, weather, and other leaves, leading to complex backgrounds.

The subject of the study described in this paper concerns the classification of apple leaf disease images acquired in the field. This study focuses on two of the most common apple leaf diseases called cedar rust and scab. Fig. 1 shows sample images of apple leaves from the PlantVillage dataset (laboratory images) and PlantPathology dataset (field images). Cedar apple rust, which results from the fungus Gymnosporangium juniperi-virginianae, manifests as yellow to orange round spots on the upper surfaces of leaves, while apple scab, caused by the fungus Venturia inaequalis, is characterized by gray to brown lesions on the leaf surfaces.

Classifying images captured in the field presents greater challenges compared to those taken in the laboratory, primarily due to the variability in background conditions, including the presence of other objects such as soil, additional leaves, and various environmental elements. Complex backgrounds in images captured under real field conditions have varying impacts on the accuracy of plant leaf disease detection and classification, and the improvement in classification performance is highly dependent on the image segmentation technique used for background removal [6, 7]. These findings stem from studies validated on various image datasets that may not fully represent all scenarios, even though they consist of plant leaf images from different species and various conditions of capture. This inherent limitation in dataset diversity could introduce bias into the results. Therefore, further exploration of the benefits of background removal for plant leaf disease classification warrants more attention from researchers.

Barbedo [6] provided an analysis of the factors that limit the performance of plant disease classification using deep learning models. The most impactful factors were categorized as extrinsic factors (background and capture conditions) or intrinsic factors (symptoms features like color, size and location). The study by Fenu and Malloci [7] highlighted the significant drop in performance from laboratory-acquired datasets to field-collected datasets and identified the background variability as the most impactful factor. An assessment of the impact of data augmentation on the performance of deep learning-based methods for plant disease detection was performed by Arsenovic et al. [8] using both traditional methods and deep learning approaches. It was reported that the models performance was better when the models were trained using both images from the real-life conditions and synthetic images generated with the generative adversarial networks.

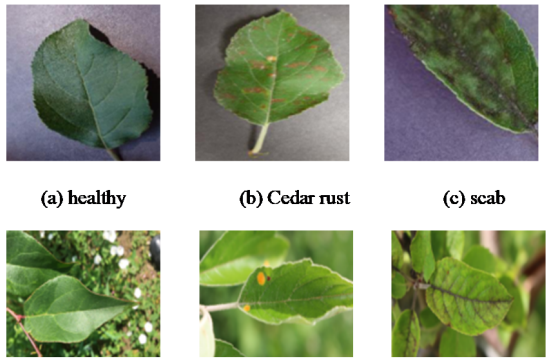

Fig.1. Sample apple leaves. Top: images of apple leaves with controlled background (PlantVillage dataset). Bottom: images of apple leaves with real-world background (PlantPathology dataset).

Background removal is the process by which the background of an image is suppressed while preserving the foreground. The foreground refers to the main object(s) of interest (plant leaf in this case) which may be healthy or diseased, while the background corresponds to the remaining area of the image. Removing the background essentially involves extracting the foreground. Background removal and foreground extraction are closely related techniques in image processing and can be regarded as a special case of image segmentation. Previously proposed methods for detecting and classifying diseases in apple leaves using machine learning techniques and deep learning models have typically been trained on datasets of leaf images captured under controlled or real-world conditions. Some methods include a background removal step to address the challenge of detecting and classifying leaf images with complex backgrounds. In contrast to backgrounds of leaf images captured under controlled conditions, leaf images acquired in fields may contain, besides the objects of interest, unwanted objects like other leaves, soil, stones, and stems. Kc et al. [9] described a method combining edge-based segmentation, background subtraction, and transfer learning of CNN models for classification of leaf images acquired under field conditions. This study demonstrated that training DenseNet121, InceptionV3, and VGG19 models on images with removed background enhanced accuracy by about 12 % on the Plant Village dataset. In their study, Karlekar and Seal [10] proposed a classification model for soybean leaf diseases. They began by converting RGB color soybean leaf images to the CIELab color space and then employed the traditional k-means clustering algorithm to eliminate the background. Subsequently, they applied a CNN model named SoyNet to the segmented images for disease classification, achieving an accuracy of 98.14%. An image segmentation method based on Chan–Vese model and Sobel operator was described by Wang et al. [11]. The background was removed based on the understanding that leaf areas are typically green, while background regions are generally non-green. This method successfully segmented overlapping leaves in images of cucumber leaves. Chuanlei et al. [12] considered three diseases in 90 apple leaf images, namely, powdery mildew, mosaic and rust. The background was first removed based on specific threshold value, and then image processing techniques and support vector machine classifier were applied. A detection accuracy of over 90% was achieved. Background removal has also been performed using deep learning models. Ngugi et al. [13] described a modified U-Net model (called KijaniNet) for isolating tomato leaf images from complex backgrounds. KijaniNet was developed by modifying the encoder stages of the U-Net model based on multi-scale feature extraction. Compared to U-Net and SegNet, KijaniNet demonstrated superior performance, achieving an mwIoU of 0.9766 and an mBFScore of 0.9439. Plant leaf segmentation based on the GrabCut algorithm [14] has been widely used to address the challenge of complex backgrounds in classification tasks. The study proposed by Bukhari et al. [15] compared three image segmentation techniques—Watershed, GrabCut, and U2-Net—for background removal from wheat stripe rust data. ResNet-18 model was trained on the resulting segmented data. The dataset segmented using U2-Net achieved the highest classification accuracy of 96.196%. Zhang et al. [16] proposed a modified Grabcut algorithm to remove most of background of weed images captured in the field. Qi et al. [17] presented a lightweight plant disease classification model that integrates GrabCut, a novel coordinate attention mechanism, and channel pruning algorithms. The model employs channel pruning in order to substantially reduce model size and computational complexity by 85.19% and 92.15%, respectively. This reduction enables the network to fulfill the deployment needs of platforms with limited storage and computational power. Lian et al. [18] proposed an improved GrabCut algorithm, called C-GrabCut, to remove backgrounds. Five pre-trained models, namely Vgg16, ResNet50, EfficientNetB0, EfficientNetB4, and EfficientNetB7, were used to classify apple leaf images from Plant Pathology dataset into four classes: rust, scab, multiply, and healthy. The C-Grabcut algorithm reduces training time from 153 to 73 seconds per epoch, and the EfficientNetB4 model demonstrates superior performance with an average accuracy of 98% and a Kappa value of 0.98. Özden [19] merged the PlantVillage dataset with the PlantPathology dataset and applied the GrabCut algorithm for background removal. The MobileNetV2 was fine-tuned using different optimizers. The study revealed that the Adagrad optimizer achieved the highest test accuracy of 91%. Luo et al. [20] proposed an improved ResNet model by improving the information flow to detect apple leaf diseases. Their enhanced ResNet model attained an accuracy of 94.99% on a field-collected dataset comprising five apple leaf diseases and healthy leaves.

Most prior works have focused on identifying leaf diseases using images captured in laboratory settings, where high performance was often achieved. However, methods that targeted the detection and classification of leaf diseases in field images have mostly shown lower performance. To assess the impact of background removal on the performance of deep learning models employed in classifying apple leaf diseases in images captured under real field conditions, we proposed in this paper the use of a binary segmentation technique to extract the foregrounds (leaf images) from training dataset images. Background-removed images were added to the raw images of the training dataset. A fine-tuned pre-trained model was then trained on this augmented training dataset. This process results in an enhanced diversity within the training dataset, which in turn improves the model's ability to learn image features that are specific to each class. The remainder of the paper is structured as follows: Section II introduces the dataset used and describes the proposed method. Section III summarizes the results obtained. The final Section concludes the paper with limitations and future work.

## II. MATERIALS AND METHODS

Since training deep learning models from scratch may require a large amount of data which may not always be available, two techniques are commonly used to significantly reduce the need for large amount of data: transfer learning and data augmentation. These two techniques were employed in this study to perform apple leaf disease classification using small datasets. The data augmentation method is based on background removal, and the classification model was chosen based on its performance, the availability of computational resources, and the intended application of the trained model.

### A. Proposed method

We propose a two-stage method for classifying apple leaf diseases in images acquired under field conditions. In the first stage, we utilized a binary segmentation technique based on deep learning to remove the background, thus isolating the healthy or diseased leaf from the original image with a cluttered background. The binary segmentation technique was solely employed on the training dataset as the goal of this study is to utilize background removal for the purpose of data augmentation. In the second stage, a pre-trained deep learning model with lightweight architecture was first fine-tuned and then utilized to assess the effect of background removal on classification accuracy.

### B. Dataset

The apple leaf dataset used in our experiments is a subset of the PlantPatholoy dataset [5]. This subset was made available to the Kaggle community platform for the PlantPathology Challenge competition as a part of the FGVC7 workshop at CVPR 2020. It contains 1821 expert-annotated RGB images categorized into four classes that are healthy, rust, scab, healthy, and multiple diseases. The images were captured under field conditions. This dataset is imbalanced since the number of images in each class is as follows: 516 healthy images, 622 rust images, 592 scab images, and 91 multiple disease images. Training deep learning models on imbalanced datasets tend to be biased towards the majority classes, which subsequently leads to poor generalization. Two primary strategies may be considered to address the class imbalance in the image dataset. In the first strategy, the number of images in each class is reduced to match the size of the smallest class. This is achieved by removing excess images from each class [7]. In the second strategy, the number of images in each class is increased to match the size of the largest class. This can performed by generating synthetic images through transformations like rotation, flipping, and scaling [18]. In our study, we did not consider the underrepresented class of multiple diseases, and adopted the first strategy by randomly selecting 516 images for classes "scab" and "rust". The resulting dataset that contains 1548 images was then divided into training, validation, and test datasets in a ratio of 6:2:2, respectively. Subsequently, the training data set was expanded using images produced by background removal in order to evaluate the effect of this type of data augmentation on the classification performance of the deep learning model.

### C. Background removal

As previously mentioned in the Introduction, background removal may be regarded as a special case of image segmentation where the image is divided into two regions: background and foreground. To achieve this binary segmentation, we used a binary image segmentation technique based a deep learning model developed for salient object detection [21]. This technique, based on the U2-Net architecture, demonstrated higher performance as compared to Watershed and GrabCut image segmentation techniques [15]. Fig. 2 shows an example of background removal. Correctly segmented images are shown on the left while incorrectly segmented are shown on the right.

### D. Deep learning model selection and transfer learning

Transfer learning involves taking a pre-trained model that was originally trained on large scale dataset and updating its parameters (weights) on a small, new dataset for a related but different task. In a closely related study, Pradhan et al. [22] compared ten pre-trained CNN models for classifying apple leaf diseases using images from the PlantVillage dataset. DenseNet201 was found to outperform the other models, achieving an accuracy of 98.75%. In the proposed study, the pre-trained deep learning model MobileNetV2 was selected because it is more suitable for limited computational resources due to its lightweight architecture [23] while achieving state-of-the-art performance for classification tasks [7, 19, 24, 25, 26]. The MobileNetV2 model was originally trained on the ImageNet dataset, which consists of 1000 classes. For the specific task of classifying apple leaf diseases from the PlantPathology dataset, the final dense layer was changed to three nodes according to the number of classes considered in this dataset. The early layers in the base model were frozen to prevent weight updates during training. The last layers in the base model extract more specialized features. Unfreezing the last layers allows its weights to be updated during the training process, thus enabling the model to learn features that are specific to the images in the dataset used in the study. A custom head model was added on top of the base model. The custom head model consists of the following layers: global average pooling, batch normalization, two dense layers with 128 and 64 nodes each using ReLU activation function, two dropout layers with a rate of 0.5, and an output layer with three classes, followed by a softmax activation function. The global average pooling layer calculates the average value of each feature map and produces a vector whose dimension equals the number of feature maps in the preceding convolutional layer. In the case of MobileNetV2, the feature maps are represented as a 3D tensor of shape 7x7x1280 and a 1D vector of size 1x1280 before and after pooling, respectively. This results in a compact representation of the extracted features, aiding in preventing overfitting and reducing computational complexity. The batch normalization layer is used to improve training speed, stability, and generalization to unseen data. The dropout layer with a rate of 0.5 provides a balance between preventing overfitting and allowing the model to learn meaningful representations from the data. The proposed framework is shown in Fig. 3.

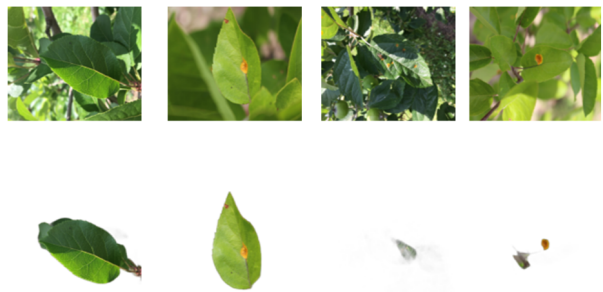

Fig. 2. Example of background removal. Top: original images. Bottom: Segmented images. Correctly segmented images (left). Incorrectly segmented images (right).

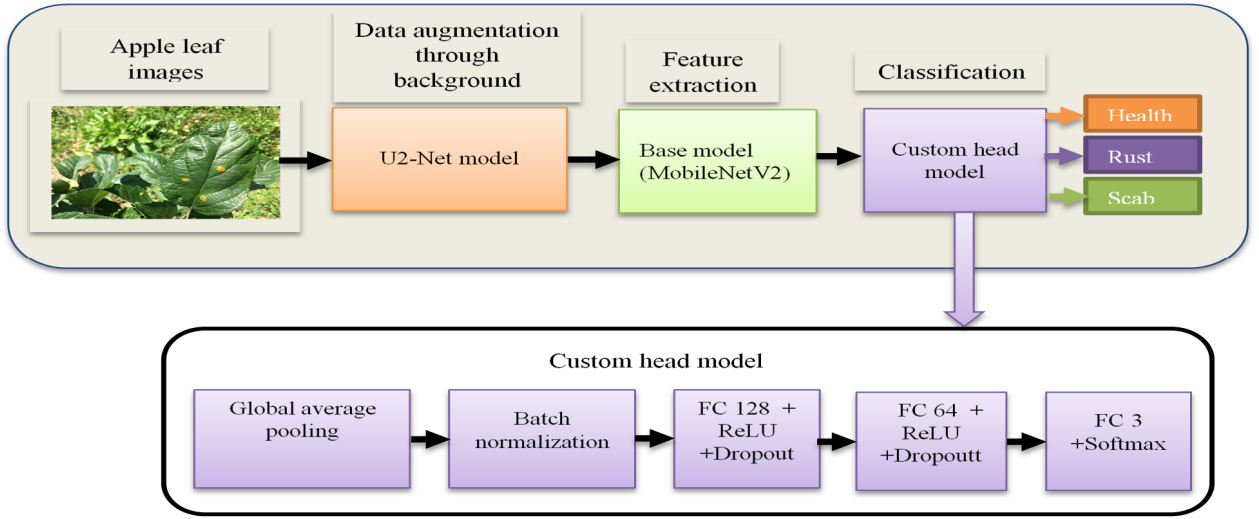

Fig. 3. Proposed method. FC: fully connected (layer).

*E. Image preprocessing and experimental setup*

The dataset used in our experiments is a subset of the PlantPathology dataset which includes three classes of apple leaf diseases: healthy, rust, and scab. Prior to data augmentation, there are 1,548 images, which are split into training, validation, and test datasets in a 6:2:2 ratio. The binary segmentation model, U2-Net, was applied to the training dataset. 18 out of 930 images were incorrectly segmented. The number of incorrectly segmented images per class is given in Table I. The pixel intensity values were normalized to be within the range [0, 1]. The dataset images were resized from their original dimensions of 2048x1365 to 224x224 to meet the input size requirements of MobileNetV2. The modified model was implemented in Python 3.9 on Windows 10 operating system using the Keras package with Tensorflow backend. The computer configuration is as follows: Intel(R) Core(TM) i5-6200U CPU at 2.30 GHz, 8 GB of RAM, GPU Intel(R)HD Graphics 520. All experiments were conducted for 50 epochs with a batch size of 16 using the categorical cross-entropy as the loss function. The modified model was trained on two training datasets using two optimizers: Adam and RMSProp. The first training dataset, denoted as dataset_1, contains 930 raw images from the PlantPathology dataset in the training dataset. The second training dataset, named dataset_2, is the first dataset_1 expanded to 1883 images with background-removed images. The images incorrectly segmented were excluded from the training dataset, resulting in a slight improvement in the classification accuracy. Adam and RMSProp optimizers were applied with their parameters given in Table II.

TABLE I. SEGMENTATION RESULTS

| Class | Number of images in each class | Incorrectly segmented images |
|---|---|---|
| Healthy | 310 | 7 |
| Rust | 310 | 8 |
| Scab | 310 | 3 |

TABLE II. PARAMETERS OF THE OPTIMIZERS

| Optimizer | Parameters |
|---|---|
| Adam | learning_rate=2e-5, beta_1=0.9, beta_2=0.99, epsilon=1e-8, amsgrad=False |
| RMSProp | learning_rate = 2e-5, rho = 0.98, epsilon = 1e-09, momentum = 0.2 |

## III. RESULTS

To show the effectiveness of the proposed method, some experiments are reported and the results obtained are compared with those of some existing methods. The metrics used to evaluate the model performance are accuracy, precision, recall, and F1-score. These metrics are computed from the confusion matrix according to their formula given in eq. (1) to eq. (4).

$$Accuracy = \frac{TP+TN}{TP+TN+FP+FN} \qquad (1)$$

$$Precision = \frac{TP}{TP+FP} \qquad (2)$$

$$Recall\ (sensitivity) = \frac{TP}{TP+FN} \qquad (3)$$

$$F1-Score = 2 \times \frac{Precision \times Recall}{Precision+Recall} \qquad (4)$$

where TP, TN, FP, and FN stand for true positive, true negative, false positive and false negative, respectively.

The modified model was first trained on the dataset_1 with Adam optimizer and then with RMSProp optimizer. In the other experiment, the modified model was first trained on the dataset_2 with Adam optimizer and then with RMSProp optimizer. The results of the evaluation of the trained model on the test datasets are presented in Tables III and IV. From these tables, it can be observed that the Adam optimizer outperforms the RMSProp optimizer in both test datasets. The confusion matrices obtained in the four cases of model training are given in Fig.4. The largest number of misclassified images belonged to the "scab class", which the model predicted to be healthy images. The model trained on dataset_2 with Adam optimizer provided the lowest number of misclassification of only 2 out of 103 scab images predicted as healthy images while the model trained on the dataset_1 with RMSProp misclassified 10 out of 103 scab images. Confusion between scab images and healthy images may be explained by the fact that scab images with less severe lesions appear like healthy images. This issue of the high similarity of the features of the images from different classes was best resolved with the proposed method.

A number of methods for classifying apple leaf diseases have been proposed in the literature. However, the results of these methods have not been obtained on a standard dataset. Consequently, they cannot be directly compared. Very few of these methods have made use of the PlantPathology dataset that was released in 2020 [5]. We report in Table V the performance of the most recent, most relevant, and highest performing works identified in the literature. From Table V, it is clear that the proposed method outperforms existing state-of-the-art works in terms of the evaluation metrics used on the PlantPathology dataset.

TABLE III. RESULTS WITH TEST DATASET_1 (RAW IMAGES ONLY)

| Optimizer | Accuracy | Precision | Recall | F1-Score |
|---|---|---|---|---|
| Adam | 95.79% | 95.98% | 95.79% | 95.81% |
| RMSProp | 94.82% | 95.22% | 94.82% | 94.80% |

TABLE IV. RESULTS WITH TEST DATASET_2 (RAW IMAGES + IMAGES WITH BACKGROUND REMOVED)

| Optimizer | Accuracy | Precision | Recall | F1-Score |
|---|---|---|---|---|
| Adam | 98.71% | 98.72% | 98.71% | 98.71% |
| RMSProp | 97.09% | 97.27% | 97.09% | 97.10% |

IV. CONCLUSION

This study investigated the impact of data augmentation using background removal on the classification performance of apple leaf disease. MobileNetV2 model was fine-tuned and trained on two field-collected datasets: one with the original images only and the other augmented by images with background removed. In addition to enhancing the dataset diversity, this type of data augmentation can help the model better learn features that are specific to each class. The trained model was evaluated on the test dataset, achieving a performance of 98.71% across all evaluation metrics used. Given its high classification performance on field-acquired datasets and its lightweight architecture, the modified MobileNetV2 trained on a dataset augmented through background removal has the potential to be transformed into a mobile application. This application would assist in apple leaf detection and classification, which is the intended practical application of this study.

TABLE V. COMPARISON OF THE PROPOSED METHOD WITH EXISTING WORKS

| Ref. | Number of classes | Accuracy | Precision | Recall | F1-Score |
|---|---|---|---|---|---|
| [7] | 3 | 87.31% | 89.24% | 87.31% | 86.59% |
| [19] | 4 | 91% | - | - | 91% |
| [18] | 4 | 98% | 98% | 98% | 98% |
| [20] | 6 | 94.23% | 94.75% | 94.23% | 94.49% |
| Proposed method | 3 | 98.71% | 98.72% | 98.71% | 98.71% |

The proposed study focused on two common diseases of apple leaves using one binary segmentation method only. It should be noted that other diseases and multiple diseases on the same leaf exist in apple leaves. The approach proposed in this work can be applied to deal with the broader challenge of disease classification in plant leaf images with complex backgrounds, utilizing any efficient binary segmentation technique. Our findings suggest that cleaned images (images without complex backgrounds) help the model learn more effectively, thereby improving its ability to accurately recognize diseases in noisy images. This observation warrants further investigation, which will be the focus of our future work.

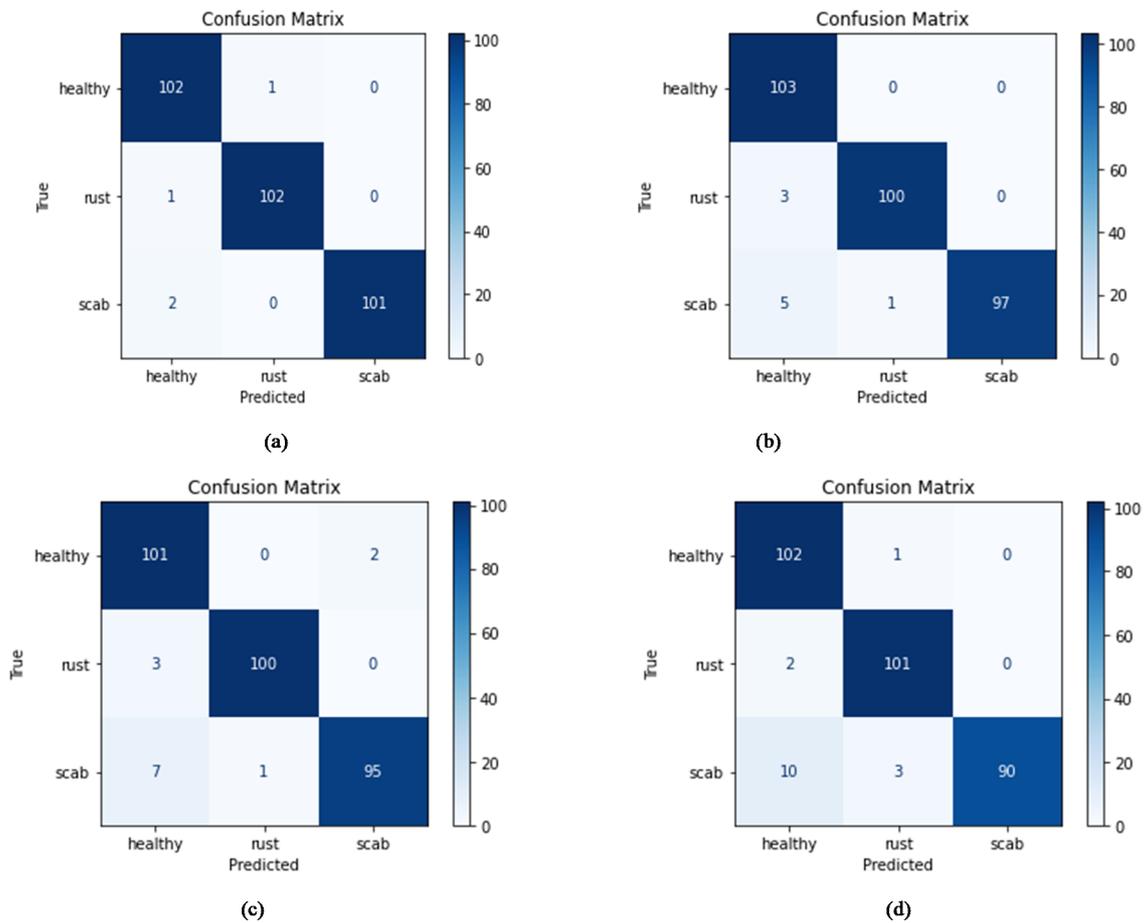

Fig. 4. Confusion matrices. (a): augmented dataset with Adam optimizer. (b): augmented dataset with RMSProp optimizer. (c): original dataset with Adam optimizer. (d): original dataset with RMSProp optimizer.